\documentclass[a4paper,12pt]{article}

\usepackage{fancyhdr}
\usepackage{fancybox}
\usepackage{amsmath}

\usepackage{graphicx}
\usepackage{subfigure}
\usepackage{multirow, multicol}

\usepackage{url}
\usepackage{amsbsy,amssymb}
\usepackage{amsmath}
\usepackage{abstract}
\usepackage{hyperref}
\usepackage{color}
\usepackage{authblk}
\usepackage{ulem}
\usepackage{fancyvrb}

\usepackage{fancyvrb}
\usepackage{multicol, tabularx}
\usepackage{changepage}
\usepackage{graphicx,psfrag,epsf}
\usepackage{longtable}
\usepackage{adjustbox}
\usepackage{lscape}

\title{Hypercomplex neural network in time series forecasting of stock data}

\author[1]{Radosław Kycia}
\author[2]{Agnieszka Niemczynowicz}

\affil[1]{Faculty of Computer Science and Telecommunications, Cracow University of Technology, Poland}
\affil[2]{Faculty of Mathematics and Computer Science,
    University~of~Warmia and Mazury in Olsztyn, Poland}

\begin{document}
\maketitle

\begin{abstract}
\noindent
The goal of this paper is to test three classes of neural network (NN) architectures based on four-dimensional (4D) hypercomplex algebras for time series prediction. We evaluate different architectures, varying the input layers to include convolutional, Long Short-Term Memory (LSTM), or dense hypercomplex layers for 4D algebras. Four related Stock Market time series are used as input data, with the prediction focused on one of them. Hyperparameter optimization for each architecture class was conducted to compare the best-performing neural networks within each class. The results indicate that, in most cases, architectures with hypercomplex dense layers achieve similar Mean Absolute Error (MAE) accuracy compared to other architectures, but with significantly fewer trainable parameters. Consequently, hypercomplex neural networks demonstrate the ability to learn and process time series data faster than the other tested architectures. Additionally, it was found that the ordering of the input time series have a notable impact on effectiveness.
\end{abstract}

\vspace{5mm}


\section{Introduction}
Many important phenomena are described by time series, including physical \cite{Bou2020}, biological \cite{Bar2012}, in medicine \cite{Wis2002}, or in economy \cite{Castro2023}. There are developing many new mathematical tools for analysis of time series \cite{Vilem2022, Bor2022, Bog2023}. One notable examples are stock exchange time series. In recent years, advances in Neural Networks (NN) outperformed \cite{NNforTimeSeries} traditional modeling based on ARIMA or state-space mathematical modeling \cite{TimeSeriesBook}. The typical NN architectures are based on convolutional and recurrent layers due to their suitability to process sequences and relate nearby data points of these sequences, e.g. \cite{CNNRNNTImeSeries}.

There are numerous directions in stock forecasting with an already existing extensive literature. Our goals encompass two crucial aspects that can significantly impact time series forecasting and other applications. The first aspect involves minimizing the number of parameters by employing new architectures. Specifically, we are investigating the utilization of novel layers based on hypercomplex numbers. This architecture is rooted in a specific dimension of the hypercomplex algebra of layers, which intuitively explores the correlations within input data of this dimension. Consequently, we can address the second aspect, which involves creating an NN model capable of combining multiple interrelated time series as inputs to predict a specific time series in the future. Such a task holds significant importance not only time series forecasting, but in many various applications, as well as. In addition, since the hypercomplex layers based on 4D algebras can treat each of four input slots differently, it is essential to examine the influence of order of input time series.
In this paper, we compare convolutional, LSTM, and hypercomplex-based layers for predicting time series. We utilize 4D algebras and select four related time series from the stock exchange for comparison. Our objective is to determine which layer demonstrates better accuracy while minimizing the number of trainable parameters. Since individual layers alone are inadequate for this task, we incorporate them into a simple supporting NN architecture. Within this architecture, hyperparameters will be optimized to compare the best models within each class. This approach aims to provide insights and qualitative results on which layer can optimize future neural networks for time series prediction. Additionally, we will examine the impact of the order of data in a quadruple input.

The paper is organized as follows: The next section provides a general description of neural network layers that are useful in time series prediction. Following that, we present the dataset description and a detailed testing methodology in the subsequent section. Afterward, we discuss the testing results, and we conclude the paper with final remarks.

\section{A brief review of deep learning time series forecasting approach}
\subsection{Classical Convolutional Neural Networks}
Convolutional neural networks (CNNs) were proposed by Yann LeCun et al. in their pioneer article \cite{cnn1}. CNNs represent a prevalent architecture in the realm of image processing and computer vision \cite{cnn2}. This success of CNNs has made them find many applications in other research areas such as medicine \cite{cnn4}, biology \cite{cnn6, cnn7}, economy \cite{cnn8, cnn9}, among others. Influenced for developing their various frameworks,  e.g., VGG16 \cite{cnn3} and Residual-Net \cite{cnn2}.

The classical usage of these networks employs usally three primary types of layers: convolutional, pooling, and fully connected layers. The fundamental role of convolutional layers is to learn features from input data. This involves applying predefined filters of a specific size to the input data through the matrix-based convolution operation. Convolution entails summing element-wise products.

Pooling, on the other hand, reduces input dimensions, thereby accelerating computation and mitigating over fitting. Among the most commonly employed pooling techniques are average pooling and max pooling. These methods aggregate values through means or maximums, respectively.

After the convolutional layers have extracted relevant features, the prediction process occurs through fully connected layers, often called dense layers. These latter layers receive flattened features resulting from the preceding convolutional and pooling layers.

\subsection{Classical Recurrent Neural Networks}
\par Recurrent Neural Networks (RNNs), first proposed by Elman et al. \cite{Elman},  are a class of artificial neural networks designed for handling sequential data. They are particularly effective when the order of the input data is important and has a temporal aspect, such as time series \cite{TS}, natural language \cite{NLP}, speech \cite{Speech}, etc.
\par There are some different architectures of RNNs. According to the number of data inputs and outputs, we can distinguish a few classical such as one-to-one, one-to-many, many-to-one, and many-to-many.

\begin{figure}[h!]
\begin{center}
\includegraphics[width=140mm,height=45mm]{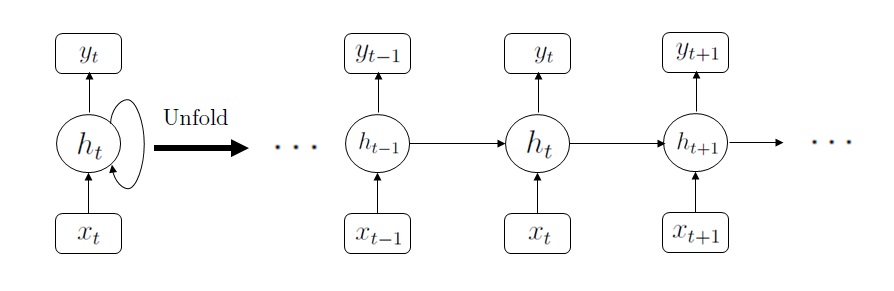}
\end{center}
\caption{Architecture RNN. On the left, there is the classical RNN structure. On the right, there is the unfolding version where the information from the previous time step ($t-1$)  is transformed to the next time step ($t$).}
\label{fig1}
\end{figure}
Unlike traditional feed-forward neural networks, where data flows in one direction from input ($x_t$) to output ($y_t$), RNNs have a feedback loop that allows information to be passed from one step of the sequence to the next (cf. Figure \ref{fig1}). This internal memory or hidden state ($h_t$) makes RNNs suitable for tasks where the current input's ($x_t$) interpretation depends on previous inputs ($x_{t-1}$) in the sequence.

\par However, classical RNNs have a drawback known as the ``vanishing gradient'' problem \cite{grad1, grad2}. When gradients (used to update the network's weights during training) become very small due to the long sequences, it becomes difficult for the network to learn dependencies that are far apart in time. This limitation led to developing more advanced RNN variants, such as Long Short-Term Memory (LSTM) networks and Gated Recurrent Units (GRUs).

\subsection{Classical Long Short-Term Memory}

Long short-term memory (LSTM) recurrent networks incorporate specialized memory cells within the recurrent hidden layer that can learn to maintain information over long periods of time \cite{LSTM-1}. They have mechanisms to control when to forget and when to update the cell state, allowing them to capture long-range dependencies in sequences.
\par  The memory blocks contains memory cells with self-connections that retain the network's temporal state alongside distinct multiplicative units known as gates, which regulate the information flow. In the classical architecture of LSTM, each memory block included an input gate and an output gate. The input gate governs the influx of input activations into the memory cell, while the output gate manages the outward flow of cell activations to the broader network. Next, the forget gate was introduced to the memory block, addressing a limitation of LSTM models that hindered their ability to process continuous input streams devoid of segmentation into subsequences. The forget gate scales the internal cell state before incorporating it as input through the cell's self-recurrent connection. This adaptive process enables the cell's memory to be either forgotten or reset. Moreover, in the modern LSTM design, there are peephole connections from the internal cells to the gates within the same cell. These connections facilitate the precise timing of learning of outputs.

\subsection{Four-dimensional hypercomplex neural networks}
Using different algebraic structures than real and complex numbers for artificial NN is an old idea \cite{HypercomplexNNBook}. However, it was revisited when sufficient computing power became available for experimenting with concrete architectures.

Recent results on hypercomplex neural networks based on 4D algebras for image processing \cite{Marcos} have shown that such networks enable the minimization of the number of trainable parameters while maintaining similar accuracy in image classification compared to classical approaches.

We focus on the hyperdense layer of hypercomplex-valued convolutional neural networks \cite{Marcos} utilizing selected 4D hypercomplex algebras. The base of these algebras consists of elements ${1, i, j, k}$ with the following multiplication matrix for ${i, j, k}$:
\begin{equation}
\text{Quaternions} = \left[\begin{array}{ccc}
     -1 & k & -j \\
     -k & -1 & i \\
     j & -i & -1 \\
    \end{array}
  \right],
\end{equation}

\vspace{3mm}

\begin{equation}
\text{Coquaternions} = \left[\begin{array}{ccc}
     -1 & k & -j\\
     -k & 1 & -i\\
     j & i & 1\\
    \end{array}
  \right],
\end{equation}

\vspace{3mm}

\begin{equation}
C\ell(1,1) = \left[\begin{array}{ccc}
     1 & k & j \\
     -k & -1 & i \\
     -j & -i & 1 \\
    \end{array}
  \right], 
\end{equation}
which we will denote as $\mathbb{H}$.

The general data flow in the hyperdense $l-$th layer is as follows:
\begin{equation}
  \mathbf{y_{l}} = f(\mathbf{w_{l}x_{l}} + {\mathbf b_{l}}),
\end{equation}
where the weights $w_{l}$ are coded as an 4D hypercomplex algebra element, it means 
\begin{equation}
    {\mathbf{w_{l}}} = w_{1}^{(l)} + iw_{2}^{(l)}+jw_{3}^{(l)} + k w_{4}^{(l)}, ~ w_{l} \in \mathbb{R}.
\end{equation}  
The input vector is split in 4-tuples 
\begin{equation}
{{\mathbf x_{l}}} = x_{1}^{(l)} + i x_{2}^{(l)} + jx_{3}^{(l)} + kx_{1}^{(l)}, ~ x_{l}\in \mathbb{R},
\end{equation}
and the bias 
\begin{equation}
{\mathbf b_{l}} = b_{1}^{(l)} + ib_{2}^{(l)} + jb_{3}^{(l)}+kb_{4}^{(l)}, ~ b_{l}\in\mathbb{R}.
\end{equation}
The function $f:\mathbb{H}\rightarrow \mathbb{H}$ is an activation/discriminating functions acting component-wise, i.e. \begin{equation}
f(v_{1}+iv_{2} + jv_{3}+kv_{4}) = f(v_{1})+if(v_{2})+jf(v_{3}) + kf(v_{4})
\end{equation}
for $v_{i}\in \mathbb{R}$, $i=1,\ldots 4$. The output $\mathbf{y_{l}}$ is treated as a $4$-dimensional vector in the next layer.

\section{Experimental setup}

\subsection{Dataset}
In applications, such as the stock market, time series within a specific domain are often correlated and exert mutual influence on each other in highly nontrivial ways.

Our focus is on the value of Copper on the NYSE, using four related stock values. We chose four values to assess the applicability of hypercomplex neural networks based on four-dimensional algebras.

The selected tickers are:
\begin{itemize}
 \item{'HG=F'- High-Grade Copper - COMEX named 'Cooper',}
 \item{'FCX'- Freeport-McMoRan company that is the largest copper miner in the world, named 'FCX',}
 \item{'SCCO'- Southern Copper company, named 'SCCO',}
 \item{'USDCLP=X'- Chilean Peso to USD exchange rate, since the largest copper mines are located in Chile. It is named 'CLP'.}
\end{itemize}

\begin{figure}[htb!]
\centering
 \includegraphics[width=140mm,height=70mm]{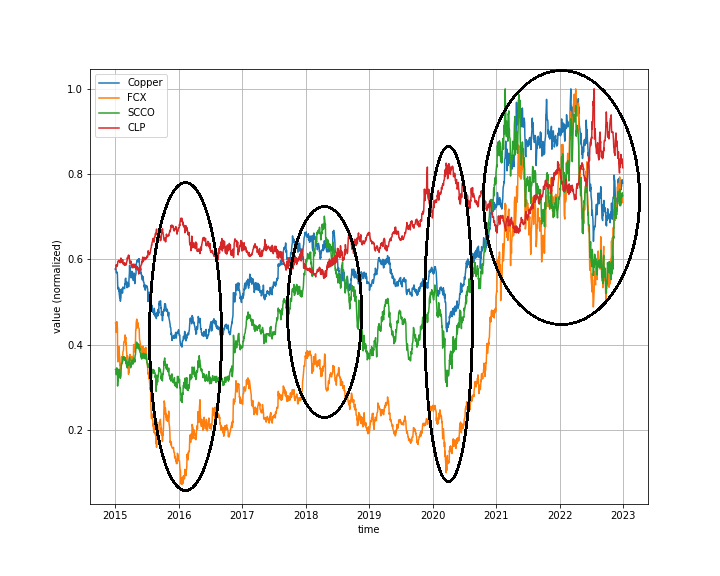}
 \caption{Selected stock prices in the period from January 1, 2015, to January 1, 2023. The ellipses marked notable correlations between values of given stock.}
  \label{Fig.StocPrices}
\end{figure}

The data were gathered within the period from January 1, 2015, to January 1, 2023, with regular sampling, occasionally interrupted due to Stock Exchange operations. They were obtained through the Yahoo! Finance service \cite{YahooFinance}. Consolidating the data into a unified data frame, any instances of missing values for a specific timestamp led to the exclusion of the entire record for that timestamp. Ultimately, the aggregated dataset comprises 2008 rows $\times$ 4 columns, as depicted in Fig.\ref{Fig.StocPrices}.

The Pearson correlation coefficients are collected in Tab. \ref{Tab.CorrelationStock} due to their common usage in analyzing linear relationships between variables. This statistical measure assesses the strength and direction of the linear association between two continuous variables, providing valuable insights into the degree of correlation between the stock values under consideration.


\begin{table}[htb!]
\begin{center}
\begin{tabular}{|c|c|c|c|c|}
\hline
 & \textbf{{Copper}} & \textbf{{FCX}} &  \textbf{{SCCO}} & \textbf{{CLP}}  \\ \hline
\textbf{{Copper}} & 1 & 0.9483 & 0.9323 & 0.4478  \\ \hline
\textbf{{FCX}} &  & 1 & 0.8682 & 0.4708  \\ \hline
\textbf{{SCCO}} &  &  & 1 & 0.3878  \\ \hline
\textbf{{CLP}} &  &  &  & 1 \\ \hline
\end{tabular}
\end{center}
\caption{Pearson correlation matrix between considered stock time series.}
\label{Tab.CorrelationStock}
\end{table}

The plots representing lagged (auto)correlations between time series for time lags between $0$ and $60$ time units are presented in Tab. \ref{Tab.AutoCorrelationStock}. One can observe strong correlation among time spans between pairs of the series. The use of lagged autocorrelation is essential in time series analysis as it helps identify any systematic patterns or dependencies between observations at different time points. This analysis aids in understanding the temporal structure of the data and can provide insights into potential forecasting models.

\begin{table}[htb!]
\begin{adjustbox}{angle=90}
\begin{tabular}{|c|c|c|c|c|}
\hline
 & \textbf{{Copper}} & \textbf{{FCX}} &  \textbf{{SCCO}} & \textbf{{CLP}}  \\ \hline
\textbf{{Copper}} & \raisebox{-\totalheight}{\includegraphics[width=0.3\textwidth]{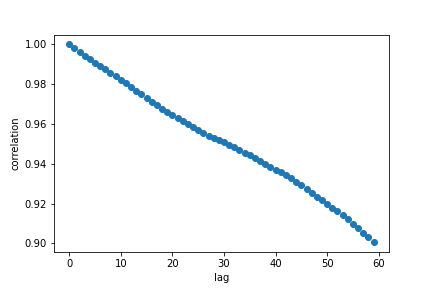}} &  \raisebox{-\totalheight}{\includegraphics[width=0.3\textwidth]{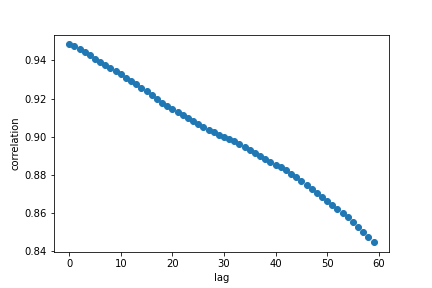}} &   \raisebox{-\totalheight}{\includegraphics[width=0.3\textwidth]{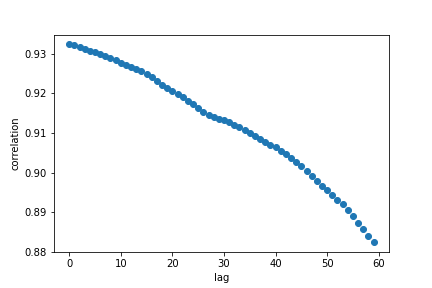}} & \raisebox{-\totalheight}{\includegraphics[width=0.3\textwidth]{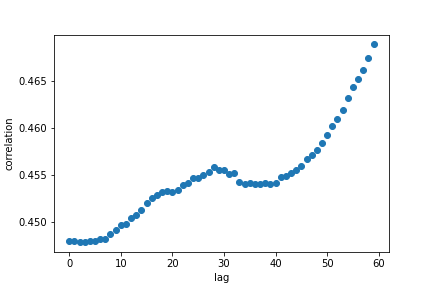}}  \\ \hline
\textbf{{FCX}} & - & \raisebox{-\totalheight}{\includegraphics[width=0.3\textwidth]{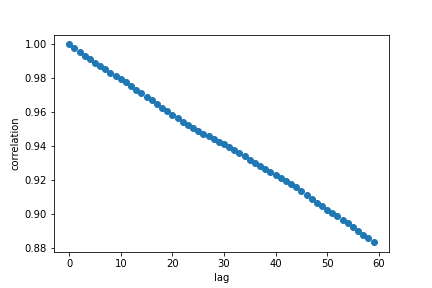}} & \raisebox{-\totalheight}{\includegraphics[width=0.3\textwidth]{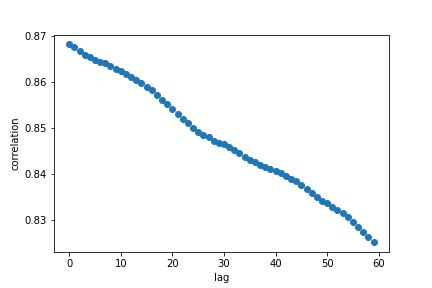}} & \raisebox{-\totalheight}{\includegraphics[width=0.3\textwidth]{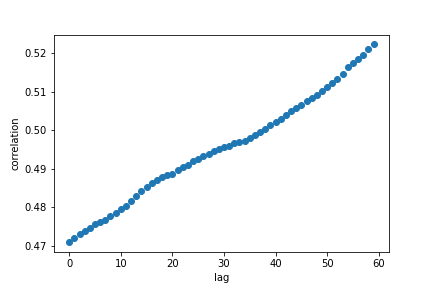}}  \\ \hline
\textbf{{SCCO}} & - & - & \raisebox{-\totalheight}{\includegraphics[width=0.3\textwidth]{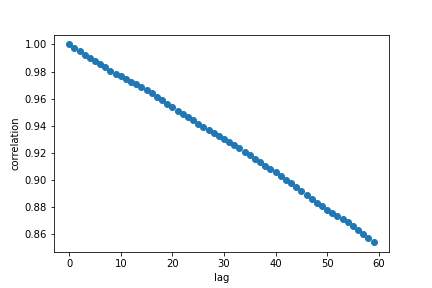}} & \raisebox{-\totalheight}{\includegraphics[width=0.3\textwidth]{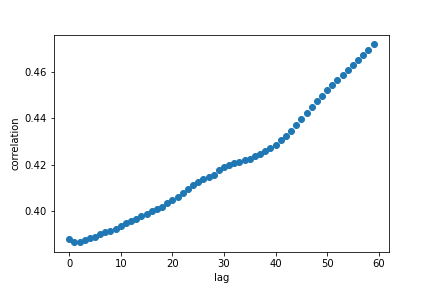}}  \\ \hline
\textbf{{CLP}} & - & - & - & \raisebox{-\totalheight}{\includegraphics[width=0.3\textwidth]{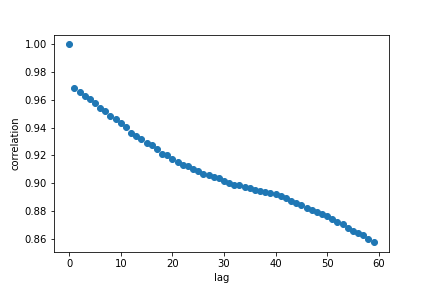}} \\ \hline
\end{tabular}
\end{adjustbox}
\caption{(Auto)correlations for pairs, with different lags, and for standarized data.}
\label{Tab.AutoCorrelationStock}
\end{table}

\subsection{Methodology}
In our experiment, our objective was to assess the predictive accuracy and computational complexity of three selected layers:
\begin{itemize}
 \item {Convolutional 1D layer from TensorFlow (Keras library),}
 \item {LSTM - Long-Short Term Memory recurrent layer form TensorFlow (Keras library),}
 \item {The Hypercomplex Dense Layer as described in \cite{Marcos}.}
\end{itemize}

\begin{figure}[htb!]
\centering
 \includegraphics[width=140mm,height=50mm]{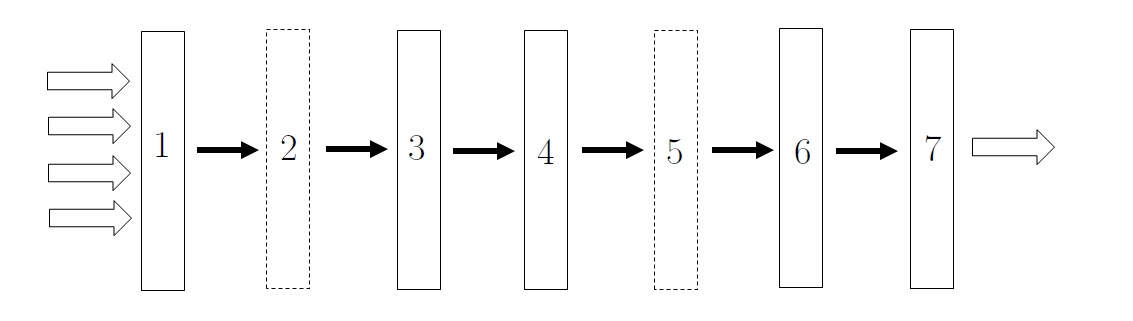}
 \caption{Testing architecture.}
  \label{Fig.TestingArchitecture}
\end{figure}

To test such layers, we need to embed them in a relatively simple general feed-forward neural network that serves as a facility for testing these layers. This whole network undergoes an optimization procedure. The general testing architecture is presented in Fig. \ref{Fig.TestingArchitecture}.
\par The architecture was optimized using grid search algorithm over a selected set of typical parameters for each layer. The layers are as follows:
\begin{enumerate}
 \item {Input layer that contains the test layer of the following type:
          \begin{itemize}
          \item {Convolutional one-dimensional layer (\texttt{keras.layers.Conv1D()} with number of filters as a parameters \texttt{n\_filters} in the range $[8, 16, 32, 64, 128]$.}
          \item {Recursive LSTM layer (\texttt{keras.layers.LSTM()}) with number of units \texttt{n\_units} in the range $[8, 16, 32, 64, 128]$.}
          \item {Hypercomplex dense layer (\texttt{HyperDense}) of \cite{Marcos} with number of units \texttt{n\_hunits} in the range $[1, 2, 4, 8, 16, 32]$ and selected $4$-dimensional algebra \texttt{algebra} from: quaternions, coquaternions, and $C\ell(1,1)$.}
         \end{itemize}
 }
 \item {Zero or one dense layer (\texttt{keras.layers.Dense}) controlled by optimized parameter \texttt{n\_dense1} from the range $[0,1]$. The number of units was optimized by the parameter \texttt{n\_units} from the range $[8,16,32,64]$. The activation function was optimized by parameter \texttt{activation} from the range $["linear", "relu"]$.}
 \item {One dimensional maximum pooling layer (\texttt{keras.layers.MaxPooling1D()}) for reducing dimensionality of the signal.}
 \item {Flattening signal (\texttt{keras.layers.Flatten()}.}
 \item {As in step 2, zero or one dense layer (\texttt{keras.layers.Dense}) is controlled by the optimized parameter \texttt{n\_dense2} from the range $[0,1]$. The number of units was optimized by the parameter \texttt{n\_units} from the range $[8,16,32,64]$. The activation function was optimized by parameter \texttt{activation} from the range $["linear", "relu"]$. We select the same parameters for layers 2 and 5 to prevent combinatorial blowup of combinations of all values of hyperparameters.}
 \item {Dropout (\texttt{keras.layers.Dropout()}) with the fraction $0.5$ units turned off from the learning process. It is a standard way for overfitting prevention.}
 \item {Dense output layer (\texttt{keras.layers.Dense()}) with a number of units adjusted to the number of time series prediction points for a given test.}
\end{enumerate}
In addition, the optimizer was set to the Adam algorithm, loss function as MSE (Mean Square Error), and metric as MAE (Mean Absolute Error). For grid search, the scoring MAE was selected as a standard measure of discrepancy between time series.

Depending on the first layer, we will call the NN testing architecture:
\begin{itemize}
 \item {CNN} - the first layer is one dimensional convolutional layer,
 \item LSTM - the first layer is LSTM layer,
 \item H - the first layer is dense hypercomplex layer.
\end{itemize}

The data were preprocessed for supervised learning requirements, resulting in the input data $X$ and the features $Y$ parts. The $X$ part contains the four time series of the fixed window length in a fixed order. The $Y$ data contains the time series to predict Cooper ticket.

We selected four cases for the \textit{window} of $X$ from the range $[10,20,40, 60]$ and four cases for predicting consecutive future values $Y$ of time series from the range $[1,5,10, 20]$ simulating short, moderate, and long prediction \textit{spans}.

After preparing the $X$, $Y$ sets, the whole data was split into $80\%$ for the cross-validation procedure and $20\%$ for validation in the NN learning procedure. With this split, we prevent data leakage during the learning process.

Then, the Grid Search algorithm was employed to optimize the space of parameters of the given architecture. In each step of the Grid Search method, the cross-validation with $10$ divisions was used.
The hyperparameter space cardinality was $159$ for CNN, $159$ for LSTM, and $575$ for H. The computations were performed on Tesla 4 GPU in the Google Colab framework.

\section{Results}
According to the optimization process, the best architecture was selected from the available points in the hyperparameter space. The best architecture within a given class of CNN, LSTM, and H was selected.

The data was arranged in the following fixed order: Copper, FCX, CLP, SCCO. The task was to predict future time series values of Copper using a neural network based on proposed architecture. This ordering is intuitive, as it distinguishes the first/real unit in the employed quaternion algebra - the ticker to predict (Cooper). The optimal models are depicted in Tab.\ref{Tab.BestArchitectures} with the designations CNN, LSTM, and H, along with details regarding the preferred algebra. 

Subsequently, the data was rearranged to: FCX, CLP, SCCO, Copper, and the optimization process was reiterated. In Tab.\ref{Tab.BestArchitectures}, the superior model was identified as HR, supplemented with specifics regarding the utilized algebra.

The table containing specific numerical values for the optimized parameters is provided in the Supplementary Materials.
\begin{table}[ht]
\begin{adjustbox}{angle=90}
\begin{tabular}{|m{1em}|c|c|c|c|c|c|}
      \hline
      & \multicolumn{5}{|c|}{input window (points)} \\\hline
       &  & 10 & 20  & 40 & 60 \\ \hline
      \multirow{6}{*}[-12em]{\rotatebox{90}{ prediction span (points)}} & 1 & \raisebox{-\totalheight}{\includegraphics[width=0.3\textwidth]{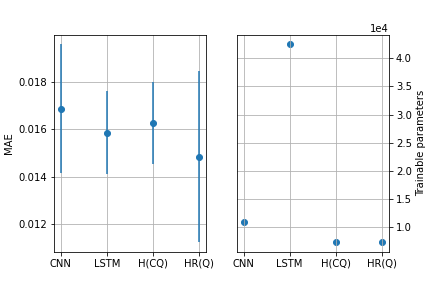}} & \raisebox{-\totalheight}{\includegraphics[width=0.3\textwidth]{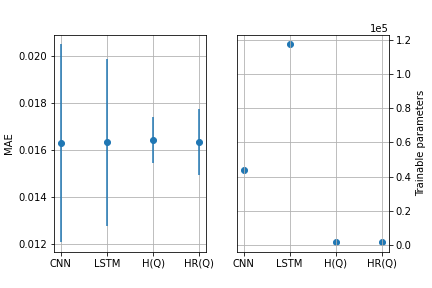}} & \raisebox{-\totalheight}{\includegraphics[width=0.3\textwidth]{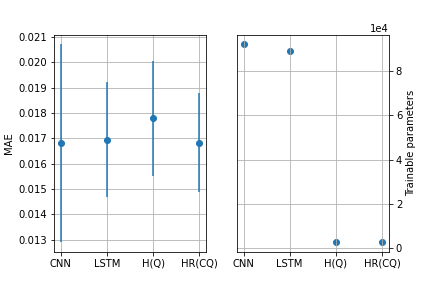}} &  \raisebox{-\totalheight}{\includegraphics[width=0.3\textwidth]{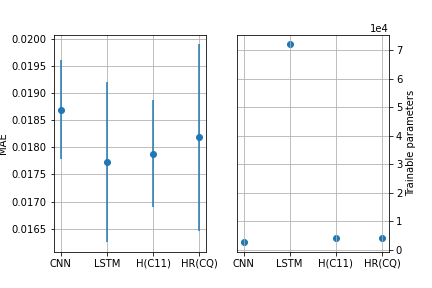}} \\ \cline{2-6}
        & 5 & \raisebox{-\totalheight}{\includegraphics[width=0.3\textwidth]{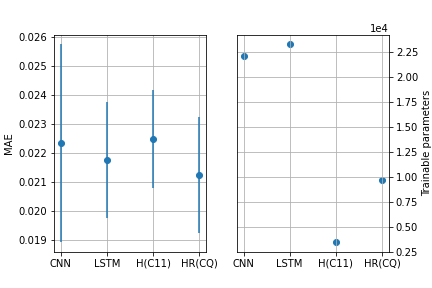}} & \raisebox{-\totalheight}{\includegraphics[width=0.3\textwidth]{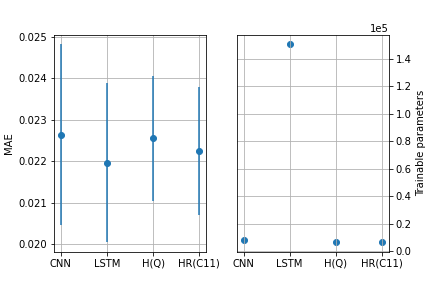}} & \raisebox{-\totalheight}{\includegraphics[width=0.3\textwidth]{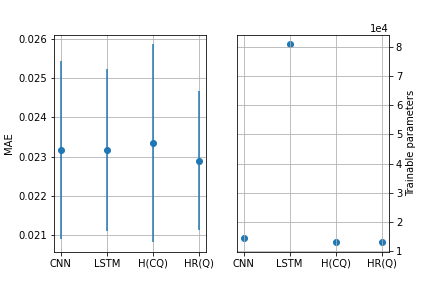}} &  \raisebox{-\totalheight}{\includegraphics[width=0.3\textwidth]{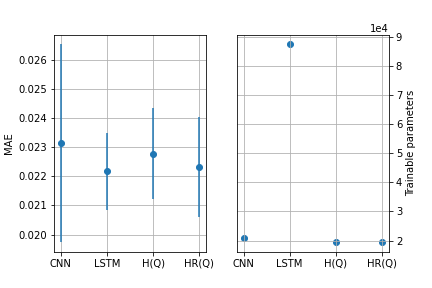}} \\ \cline{2-6}
          & 10 & \raisebox{-\totalheight}{\includegraphics[width=0.3\textwidth]{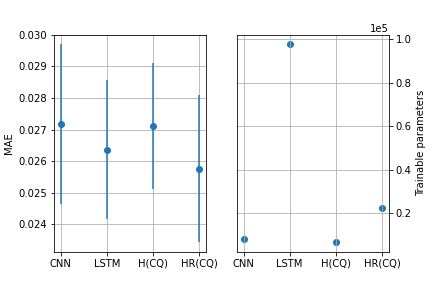}} & \raisebox{-\totalheight}{\includegraphics[width=0.3\textwidth]{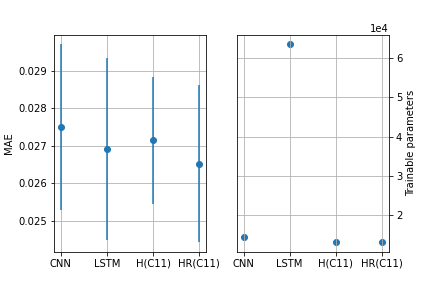}} & \raisebox{-\totalheight}{\includegraphics[width=0.3\textwidth]{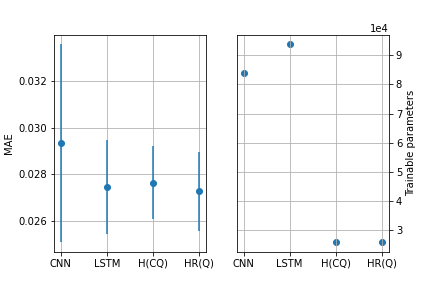}} &  \raisebox{-\totalheight}{\includegraphics[width=0.3\textwidth]{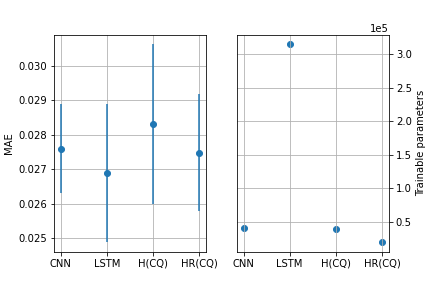}} \\ \cline{2-6}
         & 20 & \raisebox{-\totalheight}{\includegraphics[width=0.3\textwidth]{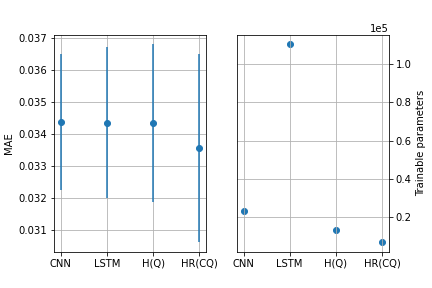}} & \raisebox{-\totalheight}{\includegraphics[width=0.3\textwidth]{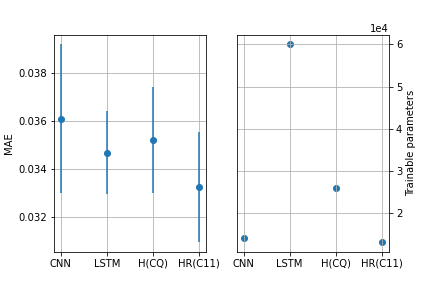}} & \raisebox{-\totalheight}{\includegraphics[width=0.3\textwidth]{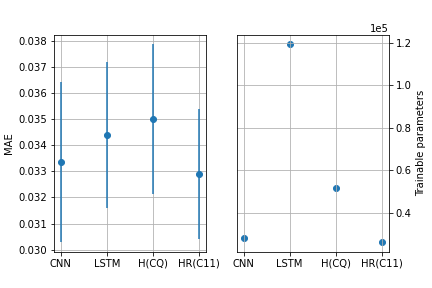}} &  \raisebox{-\totalheight}{\includegraphics[width=0.3\textwidth]{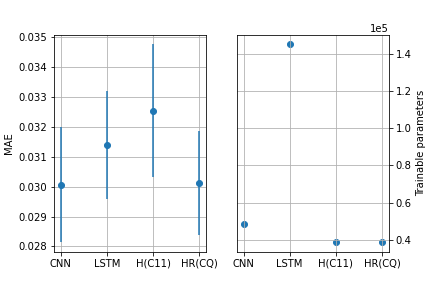}} \\ \hline
\end{tabular}
\end{adjustbox}
\caption{Scores and a number of trainable parameters for different windows and prediction spans for $CNN$, $LSTM$, and $H$ architectures for the input order data (Copper, FCX, CLP, SCCO). The label $HR$ describes the best model for the hypercomplex architecture with the input data order (FCX, CLP, SCCO, Copper).}
\label{Tab.BestArchitectures}
\end{table}

The trainable parameters encompass all weights within the optimized neural network architecture, as it is essential to include parameters from all layers for effective comparison of architecture effectiveness.

It is noteworthy that the HNN exhibits similar Mean Absolute Error (MAE) accuracy compared to other architectures; however, it typically necessitates fewer training parameters. This reduction in parameters accelerates both training and prediction, which holds significance in various applications such as stock prediction or low-power consumption device applications (e.g., in human-endpoint search engines, wearables, and IoT devices).

Furthermore, the reordering of input data can sometimes lead to a decrease in the number of trainable parameters. Thus, the assumption that the first input time series in the quadruple input data is distinguished is not consistently supported by the results. Consequently, optimization over hyperparameters and even the order of input data should be conducted to identify the optimal architecture for a specific purpose.

In our testing, we confined ourselves to four-dimensional hypercomplex algebras, and thus, models with four time series inputs were assessed. However, by utilizing higher-dimensional algebras, the results can be extrapolated to models incorporating more time series inputs.

\section{Conclusions}
We conducted tests on three architectures (CNN, LSTM, HNN) for predicting a single time series output based on quadruple time series inputs. Within each architecture class, optimization was performed. The outcomes reveal that architectures incorporating a hypercomplex dense layer, while achieving comparable MAE scores to other optimized classes, typically entail significantly less trainable parameters. This suggests that employing hypercomplex neural networks is advantageous for applications necessitating rapid learning and/or prediction times. Due to their reduced parameter count, such architectures can be deployed in constrained environments with limited resources, such as embedded systems. Additionally, the order of input data in the quadruple also influences the number of trainable parameters.

The results indicate that hypercomplex neural networks or neural networks with hypercomplex layers can offer enhanced efficiency in time series processing, akin to the well-established efficiency observed in image processing. This result is of paramount importance for specialists in time series prediction due to its implications for optimizing model efficiency and resource utilization in real-world applications.

\subsection*{Acknowledgement}
We thank Marcos E. Valle for explaining the details of hypercomplex neural networks.

This paper has been supported by the Polish National Agency for Academic Exchange Strategic Partnership Programme under Grant No. BPI/PST/2021/1/00031 (\url{nawa.gov.pl}).





\begin{thebibliography}{99}

\bibitem{HypercomplexNNBook}
P. Arena, L. Fortuna, G. Muscato, M.G. Xibilia, \emph{Neural Networks in Multidimensional Domains}, Springer-Verlag London 1998; DOI: \url{https://doi.org/10.1007/BFb0047683}

\bibitem{Bar2012}
Z. Bar-Joseph, A. Gitter, I. Simon, \emph{Studying and modelling dynamic biological processes using time-series gene expression data}, Nat Rev Genet 13, 552–564 (2012). \url{https://doi.org/10.1038/nrg3244}
\bibitem{grad1}
Y. Bengio, P. Simard, P. Frasconi, {\it Learning long-term dependencies with gradient descent is difficult} IEEE Transactions on Neural Networks, 5(2)  157--166 (1990)

\bibitem{cnn6}
P. Bhardwaj, T. Guhan, B. Tripathy, \emph{Computational Biology in the Lens of CNN} in Roy, S.S., Taguchi, YH. (eds) Handbook of Machine Learning Applications for Genomics. Studies in Big Data, vol 103. (2022) Springer, Singapore. \url{https://doi.org/10.1007/978-981-16-9158-4_5}

\bibitem{Bog2023}
B. Bichescu \& G. G. Polak, \emph{Time series modeling and forecasting by mathematical programming} Comput. Oper. Res. 151, C. (2023); \url{https://doi.org/10.1016/j.cor.2022.106079}

\bibitem{Bor2022}
J.D. Borrero, J. Mariscal, \emph{Predicting Time Series Using an Automatic New Algorithm of the Kalman Filter}, Mathematics, 10, 2915 (2022); DOI: \url{https://doi.org/10.3390/math10162915}

\bibitem{Bou2020}
N. Boull\'{e}, V. Dallas, Y. Nakatsukasa, D. Samaddar, \emph{Classification of chaotic time series with deep learning}, Physica D: Nonlinear Phenomena, 403, 132261, (2020); DOI: \url{https://doi.org/10.1016/j.physd.2019.132261.}

\bibitem{cnn9}
Ch. Chen, P. Zhang, Y. Liu, J. Liu, {\it Financial quantitative investment using convolutional neural network and deep learning technology}, Neurocomputing, {\bf 390}, 384--390 (2020); DOI: \url{https://doi.org/10.1016/j.neucom.2019.09.092}

\bibitem{Castro2023}
C.H.D. C. de Castro, F.A.L. Aiube, \emph{Forecasting inflation time series using score-driven dynamic models and combination methods: The case of Brazil}, Journal of Forecasting, 42(2), 369–401 (2023); DOI: \url{https://doi.org/10.1002/for.2908}

\bibitem{Elman}
J. L. Elman, {\it Finding structure in time} Cogn. Sci. {\bf 14}, 2 179--211 (1990)


\bibitem{cnn2}
K. He, X. Zhang, S. Ren,  J. Sun, \emph{Deep residual learning for image recognition}, in Proceedings of the IEEE conference on computer vision and pattern recognition  770--778 (2016)

\bibitem{TS}
H. Hewamalage, C. Bergmeir, K. Bandara, \emph{Recurrent Neural Networks for Time Series Forecasting: Current status and future directions}, International Journal of Forecasting, 37, 1, 388--427 (2021); DOI: \url{https://doi.org/10.1016/j.ijforecast.2020.06.008}

\bibitem{LSTM-1}
S. Hochreiter, J. Schmidhuber, \emph{Long short-term memory}, Neural Comp. {\bf 9}  8  1735 -- 1780 (1997)

\bibitem{cnn7}
J. Kim, O. Sangjun, Y. Kim, M.Lee, \emph{Convolutional neural network with biologically inspired retinal structure}, in Procedia Computer Science, 7th Annual International Conference on Biologically Inspired Cognitive Architectures, BICA 2016 {\bf 88}, 145–154 (2016); DOI: \url{https://doi.org/10.1016/j.procs.2016.07.418 (2016)}

\bibitem{NLP}
 H. Lane, H. Hapke, C. Howard, \emph{Natural Language Processing in Action}, Manning; First Edition, 2019

\bibitem{cnn1}
Y. LeCun, B. Boser, J. S. Denker, D. Henderson, R. E. Howard, W. Hubbard, L. D. Jackel,  \emph{Backpropagation Applied to Handwritten Zip Code Recognition}, Neural Comput {\bf 1} 4,  541-–551  (1989); DOI: \url{https://doi.org/10.1162/neco.1989.1.4.541}

\bibitem{NNforTimeSeries}
P. Lara-Ben\'{\i}tez, M. Carranza-Garc\'{\i}a, J.C. Riquelme, \emph{An Experimental Review on Deep Learning Architectures for Time Series Forecasting}, International Journal of Neural Systems, Vol. 31, 3 2130001 (2021); DOI: 10.1142/S0129065721300011

\bibitem{Speech}
W. Lim, D. Jang, T. Lee, \emph{Speech Emotion Recognition using Convolutional and Recurrent Neural Networks,} 2016 APSIPA IEEE, Jeju, Korea (South), 1--4 (2016); DOI:  \url{doi: 10.1109/APSIPA.2016.7820699}

\bibitem{Vilem2022}
V. Nov\'{a}k, \emph{How Mining and Summarizing Information on Time Series Can Be Formed Using Fuzzy Modeling Methods}, in: Kahraman, C., Tolga, A.C., Cevik Onar, S., Cebi, S., Oztaysi, B., Sari, I.U. (eds) Intelligent and Fuzzy Systems. INFUS 2022. Lecture Notes in Networks and Systems, vol 504. Springer, Cham. 2022; \url{https://doi.org/10.1007/978-3-031-09173-5_7}


\bibitem{grad2}
R. Pascanu, T. Mikolov, Y. Bengio,  \emph{On the difficulty of training recurrent neural networks}, in Proceedings of the 30th International Conference on Machine Learning,  1310–-1318 (2013)

\bibitem{CNNRNNTImeSeries}
K. Wang, K. Li, L. Zhou, Y. Hu, Z. Cheng, J. Liu, C. Chen, \emph{Multiple convolutional neural networks for multivariate time series prediction}, Neurocomputing, 360, 107--119 (2019)
DOI: \url{https://doi.org/10.1016/j.neucom.2019.05.023}

\bibitem{Wis2002}
Wism\"{u}ller, A., Lange, O., Dersch, D.R. et al. \emph{Cluster Analysis of Biomedical Image Time-Series}, International Journal of Computer Vision 46, 103–128 (2002); DOI: \url{https://doi.org/10.1023/A:1013550313321}

\bibitem{Marcos}
G. Vieira, M. E. Valle, \emph{Acute Lymphoblastic Leukemia 'Detection Using Hypercomplex-Valued Convolutional Neural Networks}, IJCNN (2022); arXiv: 2205.13273


\bibitem{cnn4}
D.R. Sarvamangala, R.V.  Kulkarni, \emph{Convolutional neural networks in medical image understanding: a survey}, Evol. Intel. {\bf 15} 1--22  (2022); DOI: \url{https://doi.org/10.1007/s12065-020-00540-3}

\bibitem{cnn3}
K. Simonyan, A. Zisserman, \emph{Very Deep Convolutional Networks for Large-Scale Image Recognition} in  The 3rd International Conference on Learning Representations (ICLR2015), \url{https://arxiv.org/abs/1409.1556}

\bibitem{TimeSeriesBook}
R.H. Shumway , D.S. Stofferm \emph{Time Series Analysis and Its Applications}, Springer Texts in Statistics, 2017; DOI: 10.1007/978-3-319-52452-8

\bibitem{YahooFinance}
Yahoo! Finance: \url{https://finance.yahoo.com/} (accessed: 10 January 2024)

\bibitem{cnn8}
Y. Zheng, Z. Xu, A. Xiao, \emph{Deep learning in economics: a systematic and critical review}, Artif Intell Rev {\bf 56}  9497–-9539 (2023); DOI: \url{ https://doi.org/10.1007/s10462-022-10272-8}


\end{thebibliography}
\end{document}